\documentclass[lettersize,journal]{IEEEtran}
\usepackage{amsmath,amsfonts}
\usepackage{algorithmic}
\usepackage{algorithm}
\usepackage{array}
\usepackage[caption=false,font=normalsize,labelfont=sf,textfont=sf]{subfig}
\usepackage{textcomp}
\usepackage{stfloats}
\usepackage{url}
\usepackage{verbatim}
\usepackage{graphicx}
\usepackage{cite}
\usepackage{booktabs}
\usepackage{multirow} 
\usepackage{pifont}
\newcommand{\cmark}{\ding{51}}%
\newcommand{\xmark}{\ding{55}}%
\usepackage{hyperref}
\usepackage{bm}
\hypersetup{
    colorlinks=true,
    linkcolor=blue,
    filecolor=magenta,      
    urlcolor=black,
    pdftitle={NAF-DPM},
    pdfpagemode=FullScreen,
    }
    
\begin{document}

\title{NAF-DPM: A Nonlinear Activation-Free Diffusion Probabilistic Model for Document Enhancement}

\author{Giordano~Cicchetti, and~Danilo Comminiello,~\IEEEmembership{Senior Member,~IEEE}
\thanks{Authors are with the Department of Information Engineering, Electronics and Telecommunications (DIET), Sapienza University of Rome, Italy. \\Corresponding author's email: giordano.cicchetti@uniroma1.it.}}

\maketitle


%
%
\begin{abstract}

Real-world documents may suffer various forms of degradation, often resulting in lower accuracy in optical character recognition (OCR) systems. Therefore, a crucial preprocessing step is essential to eliminate noise while preserving text and key features of documents. In this paper, we propose NAF-DPM, a novel generative framework based on a diffusion probabilistic model (DPM) designed to restore the original quality of degraded documents. While DPMs are recognized for their high-quality generated images, they are also known for their large inference time. To mitigate this problem we provide the DPM with an efficient nonlinear activation-free (NAF) network and we employ as a sampler a fast solver of ordinary differential equations, which can converge in a few iterations. 
To better preserve text characters, we introduce an additional differentiable module based on convolutional recurrent neural networks, simulating the behavior of an OCR system during training. Experiments conducted on various datasets showcase the superiority of our approach, achieving state-of-the-art performance in terms of pixel-level and perceptual similarity metrics. Furthermore, the results demonstrate a notable character error reduction made by OCR systems when transcribing real-world document images enhanced by our framework. Code and pre-trained models are available at \href{https://github.com/ispamm/NAF-DPM}{https://github.com/ispamm/NAF-DPM}.
\end{abstract}


%
%

\section{Introduction}
\IEEEPARstart{D}{ocument} images, captured by cameras, scanners, and commercial reading devices, are typically dirty and noisy. Numerous factors contribute to the low quality of document images such as low-accurate devices, bad lighting conditions, improper shooting angles, shadows or blurred text. Additional noise can be caused by the presence of stains, watermarks, ink artifacts or by background textures - for instance, colored images, logos, photocopies, and wrinkled papers. These are the reasons why degraded documents are often challenging for modern optical character recognition (OCR) systems to be transcribed. The accuracy of OCR systems is also crucial for all the downstream tasks that rely on them such as key information extraction, text mining, text-to-speech (TTS) or automatic machine translation. 
Therefore, restoring the original quality of the degraded document is of fundamental importance.

\begin{figure}[!t]
\centering
\includegraphics[width=0.9\linewidth]{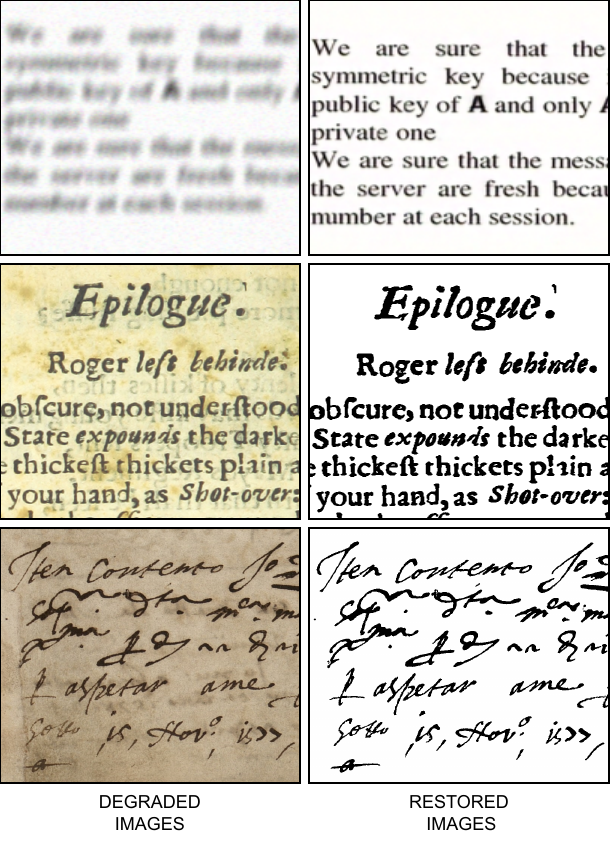}
\vspace{-0.7cm}
\caption{Random samples of image patches extracted from the Document Deblurring dataset, DIBCO2017 and H-DIBCO2018 \cite{DeblurringDataset,DIBCO2017,DIBCO2018}. The first column refers to degraded images; The second one refers to their counterparts restored by our framework. These enhanced images are high-quality and very similar to the original ones. Furthermore, the edges of the text appear to be sharper and well defined. All this benefits the work of an OCR that will be able to better recognize text.}
\label{fig_1}
\vspace{-0.7cm}
\end{figure}

One of the most popular tasks within the field of document enhancement is document deblurring. The goal of this task is to remove blur patterns and restore the original readability of the documents \cite{survey}. Document binarization is another widely studied problem, whose goal is to classify each pixel either as a background or a text pixel in order to highlight the relevant content\cite{review}. The major challenges presented by these problems are the removal of noise while keeping intact the text, the shape of every single character (both machine-printed or hand-written), and the basic features of documents \cite{challenges}. Intense noise often leads to the complete degradation of the text, and reproducing the correct text output is a daunting task. The presence of watermarks, for example, introduces a large amount of dirt, making the document difficult to read. While locating the watermark might be an affordable task, its removal and subsequent recovery of the real occulted information turns out to be very challenging. 
An additional challenge is the inference time, especially in an industrial environment where there is the need to process high-quality images in a short amount of time. This requires the development of a framework with high restoration capabilities but at the same time with low inference latency \cite{FAIR, DEGAN,docentr,DocDiff}.

In recent years, one of the most promising and effective approaches to address these challenges has involved deep generative models. 
Among them, variational autoencoders (VAEs) \cite{vae1,vae2,vae3}, generative adversarial networks (GANs)\cite{gan1,gan2,gan3}, and diffusion probabilistic models (DPMs) \cite{DDPM,DDIM,palette} have achieved great notoriety. The latter ones, DPMs, are now considered the most reliable choice for high-quality image generation problems (both conditioned and unconditioned). Indeed, it has been shown that they are surprisingly good in sample quality, surpassing GANs in image generation  \cite{diffvsgan}, and they also feature good mode coverage and diversity, as indicated by high likelihood. However, generating new items from them often requires thousands of neural network evaluations, making their application to real-world problems expensive in terms of time and computational power.

In this paper, we propose a novel generative framework based on DPMs that solves document enhancement tasks by treating them as conditional image-to-image translation problems. 
To design a specific DPM-based framework for document enhancement tasks, we focus on two main problems: the large sample diversity, which can lead to inconsistent character restoration, and the huge inference times.

To reduce the diversity of the model, we introduce a deterministic predictor before the DPM model that tries to restore as many low-frequency details as possible. 
To shorten inference time, we worked on both the architecture of the framework and the sampling strategy level. Regarding the architecture, we investigate the power and effectiveness of a nonlinear activation-free network (NAFNet) \cite{NafNet}. Such a network turned out to be very effective for the image denoising task along with a substantial reduction in computational resources needed. Motivated by these results, we design a lightweight variant of NAFNet to be used as backbone network for a DPM.

In order to guarantee a good sampling quality in a few iterations, we have adopted a deterministic sampling strategy based on an ordinary differential equation (ODE) fast solver \cite{dpmsolver}.
Experiments point out that performance converges in at most 20 iterations, instead of hundreds or even thousands of iterations required by alternative sampling strategies.
%
Finally, to make our framework robust to character substitution errors, we have included an additional differentiable module that simulates the behavior of a commercial OCR system.

We evaluate the effectiveness of our framework on two representative tasks for document enhancement: deblurring and binarization. The results demonstrate the superior performance of our framework with respect to state-of-the-art methods. Within the document deblurring task, we achieve a significant improvement on the widely used peak signal-to-noise ratio (PSNR) metric with respect to DE-GAN \cite{DEGAN} and DocDiff \cite{DocDiff}.
On the same task, we validate the strength of our framework using a commercial OCR system. Surprisingly, the images restored by our framework show a high reduction of character errors detected by OCR, halving the character error rate (CER) compared to DocDiff \cite{DocDiff}.
For document binarization we test our framework on the three most recent datasets from DIBCO Challenges \cite{DIBCO2017,DIBCO2018,DIBCO2019}. We reach competitive results for every dataset, setting a new state-of-the-art for the challenging DIBCO2019 dataset.


%
%

\section{Related Work}

Document enhancement is a vast area that includes a lot of different subtasks such as deblurring, background texture removal, page smudging, watermark removal and handwriting fading. In this section, We group some of the most relevant works in these areas into tasks: document deblurring and document binarization. The advent of diffusion models in recent years has brought an incredible increase in performance in various generative deep learning tasks. However, there are few works based on this technology in the field of document enhancement. We have devoted a section to summarize some of the remarkable work based on conditional diffusion models for the more general category image-to-image transformation.

\subsection{Document Deblurring}
\begin{figure*}[!t]
\centering
\includegraphics[width=\linewidth]{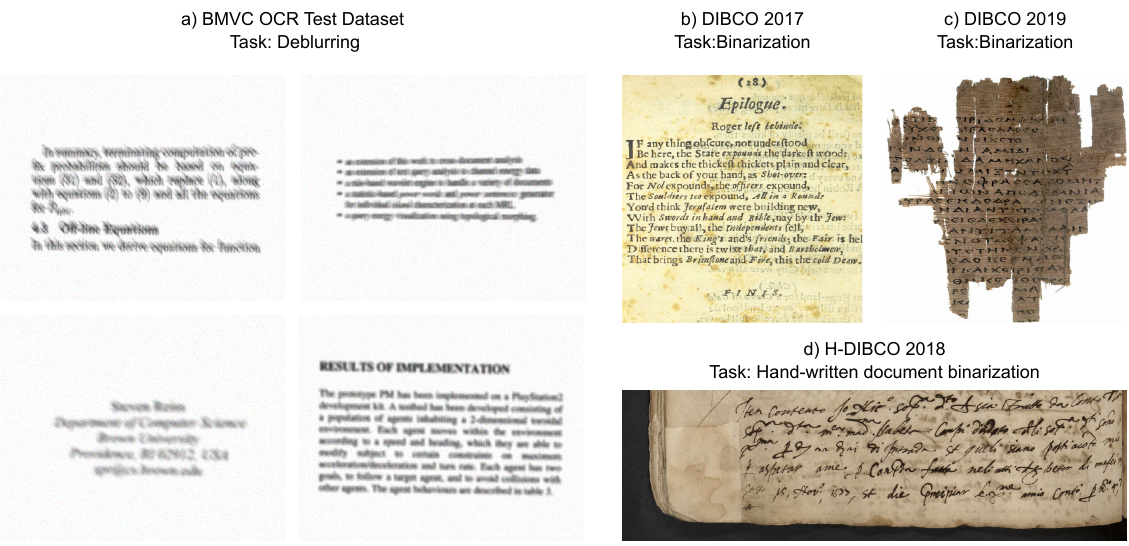}

\caption{Examples of document images used during this work. Images in subfigure (a) come from document deblurring OCR test dataset \cite{DeblurringDataset} and their associated task is image deblurring. Images in subfigures (b),(c),(d) come from the annual Document Image Binarization Competition (DIBCO) \cite{DIBCO2017,DIBCO2019,DIBCO2018}. and their associated task is image binarization.}
\label{Dataset images}
\end{figure*}

One of the most common disturbances that affect document images is blurring. Fuzzy characters can be very difficult to be recognized by an OCR system and sometimes even by the human eye (Fig. \ref{Dataset images}a). 
Before the advent of deep learning, prominent approaches against document deblurring were based on estimating the blur kernel and then removing it using non-blind deconvolution \cite{bussgang,cho2012,pan2014}.
Then in 2015 Hradis \textit{et al}. proposed a method based on convolutional neural network (CNN) \cite{DeblurringDataset}. The trained networks are able to perform blind image deconvolution without assuming any blur kernel or noise pattern.
Later on, remarkable success was made by GAN-based architectures such as DeblurGAN \cite{DEBLURGAN}, DEGAN \cite{DEGAN},  Blur2sharp \cite{Blur2Sharp} and DeblurGAN-CNN \cite{DeblurGanCNN}. 
All the aforementioned models are trained to reduce pixel-level loss ($\mathcal{L}_1$ or $\mathcal{L}_2$). Their final aim is to reach high values on distortion metrics such as peak signal-to-noise ratio (PSNR) and structural similarity index measure (SSIM). However, it has been shown that distortion metrics and perceptual quality are at odds with each other \cite{distortion}. As a result, text restored from these models has blurred and distorted edges.
Recently was presented the first diffusion model applied to document enhancement tasks: DocDiff \cite{DocDiff}. The images refined by the diffusion-based network appear to be sharper with respect to previous methods and this directly affects the performance of OCR systems. 

\subsection{Document Binarization}

Document binarization aims to classify each pixel of an image as either a background pixel or a text pixel. At the end of the process, we can obtain a binary image where only the information concerning text and important features of the image are retained. 
Different types of degradation occurring in documents can be classified as binarization problems: background texture removal, page smudging, watermark removal, handwriting fading and others (Fig. \ref{Dataset images} b,c,d).

Binarization techniques can be divided into threshold-based and numerical-based methods.
Well known threshold-based techniques are Otsu \cite{otsu}, Sauvola  \cite{SAUVOLA} and Niblack algorithms. They are able to separate the foreground from the background of an image by finding a threshold value that maximizes the variance between the two classes. Lelore \textit{et al}. \cite{FAIR} introduced a model called FAIR. It is based on edge detection to localize the
text in a degraded document image. These approaches guarantee fast computation speed and broad applicability, but, on the other hand, the binarization process is not ideal in some circumstances. 
Numerical-based methods try to classify each pixel based on the context in which they are surrounded. In 2019, Bezmaternykh \textit{et al}. proposed a model based on U-NET (encoder-decoder network with skip-connection) \cite{Bezmaternykh}. Generative adversarial networks also show their potential for this task \cite{DEGAN}. In 2022, DocEnTr was proposed. It is a transformer-based network that sets the state-of-the-art performance on different document binarization datasets \cite{docentr}.

\subsection{Document Enhancement with Diffusion Models}

The advent of Diffusion Probabilistic Models (DPMs) was a significant breakthrough in generative deep learning. It has been shown that they are surprisingly good in sample quality, beating GANs in image generation \cite{diffvsgan}. They also feature good mode coverage and diversity, indicated by high likelihood. However, generating new items from them often requires thousands of neural network evaluations, making their application to real-world problems expensive in terms of time and computational power. 
Diffusion models have been employed in all computer vision tasks:  image super resolution, inpainting, restoration, translation, and editing. Super-Resolution via Repeated Refinement (SR3) \cite{sr3} is one of the remarkable works in this field. It uses DPM to enable conditional image generation, conducting super-resolution through a stochastic iterative denoising process. Similar to SR3, Palette \cite{palette} makes use of a conditional diffusion model to create a unified framework for four image generation tasks: colorization, inpainting, uncropping, and JPEG restoration. 
Some works try to use a DPM to predict the residual image, which then can be added to the degraded one to restore its quality \cite{srdiff,debludiff,resdiff}. 
As already said, the major drawback of diffusion models is their huge inference time. To speed up the sampling phase the main techniques used are: advanced SDE/ODE solvers \cite{dpmsolver,solver2,sasolver}, distillation techniques \cite{distillation,adversarialDiffDist,ondistillation} and low-dimensional diffusion models \cite{VDM,latentdiffusion,latent2}.
The fast ODE solver from \cite{dpmsolver} is able to converge in a few iterations without sacrificing quality. 
\begin{figure*}[!t]
\centering
\includegraphics[width=7in]{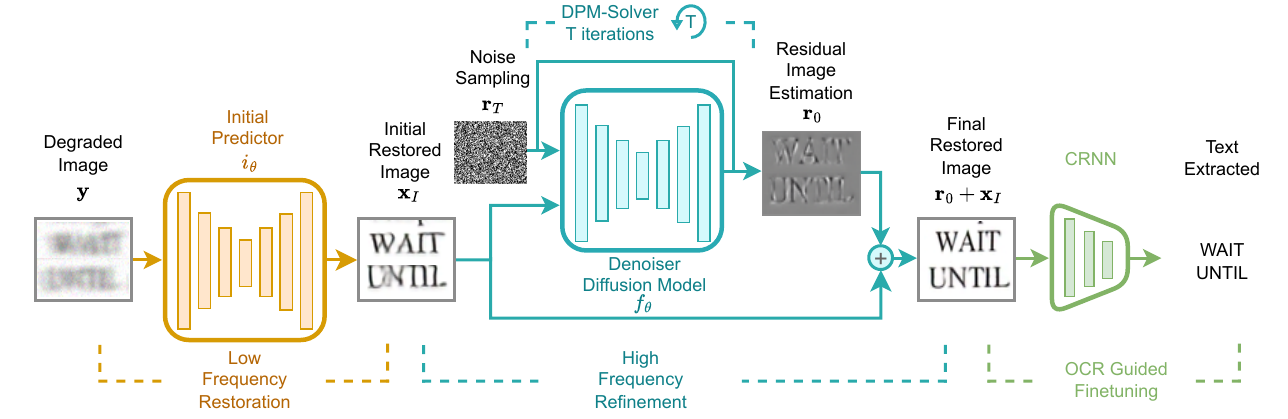}
\caption{NAF-DPM architecture. An initial predictor retrieves the low-frequency information, and then a denoiser network estimates the residual high-frequency details by iterative refinement. The high-frequency information is restored estimating the residual image that, at the end, is added back to the image predicted by the initial predictor network. As the backbone network for the initial predictor we employ an efficient nonlinear activation-free network (NAFNet \cite{NafNet}). As the backbone network for the denoiser diffusion model we design a novel and effective variation of NAFNet that takes into consideration the conditioning of the timestep $t$ of diffusion models in order to improve performance in denoising and deblurring tasks. It is worth noting that the initial predictor significantly retrieves the structure of the document while the denoiser restores all the details that make the text correct and readable. To better preserve text characters, we finetune our framework with
an additional differentiable module based on convolutional recurrent neural networks, simulating the behavior of an OCR system during training }
\label{architecture}
\end{figure*}

%
%

\section{Proposed Approach}

\subsection{Overview Of The Proposed NAF-DPM}
\label{ddpm}
In our work we propose a generative framework based on a nonlinear activation-free diffusion probabilistic model (NAF-DPM). It exploits the idea ``predict-and-refine" introduced in \cite{debludiff} and further refined in \cite{resdiff,DocDiff}. As illustrated in Fig.~\ref{architecture}, we first employ an initial predictor that takes the degraded images and produces an enhanced version by removing as much noise as possible. Then a DPM network only needs to restore the residual details.
The two networks are jointly trained and a frequency separation technique is employed. The initial predictor is responsible for the low-frequency information restoration, while the denoiser is responsible for the high-frequency information restoration. Various studies have shown that this approach is able to effectively generate images with much more defined details \cite{resdiff,freqSR,fresep2,DocDiff}. To separate the frequencies of an image we use simple linear filters, similar to \cite{resdiff,DocDiff}. Using a low-pass filter $\phi_L$ and a high-pass filter $\phi_H$ we can rewrite an image $\textbf{x}$ using:
\begin{equation}
    \mathbf{\bar{x}} = \phi_L \ast \mathbf{x} + \phi_H \ast \mathbf{x}
    \label{frequency_separation}
    \vspace{0.1cm}
\end{equation}
To better preserve text characters and to teach the network how to restore all the details, we finetune our framework with an additional differentiable module based on convolutional recurrent neural networks, simulating the behavior of an OCR system during training.  This module is used to extract text from reconstructed images and compute an additional loss function that better guides the diffusion model to restore the original quality of the text in document images.

\subsection{Initial predictor}
The initial predictor takes the degraded image and retrieves as much low-frequency information as possible. We use a scaled-down version of a nonlinear activation-free network (NAFNet), since it has demonstrated remarkable performance in tasks such as denoising and deblurring \cite{NafNet}. NAFNet takes inspiration both from a U-Net, for extra-block organization, and from a Vision Transformer, for intra-block organization. All the components are well designed to be efficient (in terms of inference time and parameters) and effective (SOTA performance). 
The initial predictor takes degraded images $\mathbf{y}$ and produces an enhanced version $\mathbf{x}_I = i_{\theta}(\mathbf{y})$. We associate to this component two dedicated loss functions. The first one tries to minimize the mean squared error between original images and enhanced ones. The second one pushes the network to restore low-frequency information. 
\vspace{0.5cm}
\begin{equation}
    L_{\text{MSE}} = \left\|\mathbf{x}_{gt}-\mathbf{x}_{I}\right\|_2^2
\end{equation}
\begin{equation}
    L_{\text{LOW}} =   \left\|\phi_L \ast\left(\mathbf{x}_{gt}-\mathbf{x}_{I}\right)\right\|_2^2
\end{equation} 
\begin{equation}
    L_{\text{INIT}} = L_{\text{MSE}} + 2 \ast L_{\text{LOW}}
    \label{lossinit}
    \vspace{0.1cm}
\end{equation}
\begin{figure}[t]
\centering
\includegraphics[width=\linewidth]{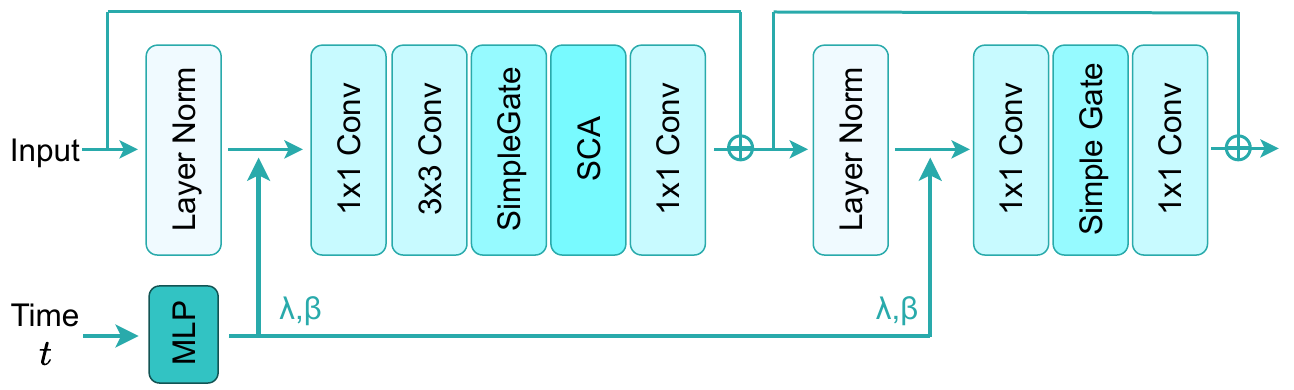}
\caption{Proposed internal structure of each NAF block of our denoiser diffusion model. We added a time processing branch, depicted in red color. This branch models the time embedding and transforms it into shift parameters $\gamma$ and $\beta$ that control the scale and bias terms of each normalization layer. SCA: Simple Channel Attention; MLP: Multi-Layer Perceptron. }
\label{modifiedNafblock}
\end{figure}

\subsection{Conditional DPM Refiner}
Diffusion models constitute a category of probabilistic generative models that systematically introduce noise to data and then learn how to reverse this process for the purpose of generating samples \cite{DDPM,DiffSurvey}.
Following the work done in \cite{debludiff,resdiff,DocDiff}, our denoiser diffusion model is trained to model the residual distribution. 

Given a ground truth image $\textbf{x}_{gt}$ and its degraded counterpart $\textbf{y}$, we define the residual image as the difference between the ground truth $\textbf{x}_{gt}$ and the enhanced image by the initial predictor $\textbf{x}_I = i_{\theta}(\textbf{y})$.
\vspace{0.1cm}
\begin{equation}
\label{residual definition}
    \textbf{r}_0 = \textbf{x}_{gt} - \textbf{x}_I
    \vspace{0.1cm}
\end{equation}
The forward diffusion process models a Markov chain that, starting from a residual sample $\mathbf{r}_0$, iteratively adds Gaussian noise according to a transition kernel $q\left(\mathbf{r}_t \mid \mathbf{r}_{t-1}\right)$ defined as follow:
\begin{equation}
    q\left(\mathbf{r}_t \mid \mathbf{r}_{t-1}\right):=\mathcal{N}\left(\mathbf{r}_t; \sqrt{\alpha_t} \mathbf{r}_{t-1}, \left( 1- \alpha_t \right)  \mathbf{I}\right),
    \label{forward_kernel}
\end{equation}
where $(\alpha_0,\alpha_1,\dots,\alpha_T)$ are hyperparameters that define the noise scheduler and they control the amount of noise that needs to be added at each timestep. According to this definition, the forward process transforms residual data distribution $ \mathbf{r}_0\sim q(\mathbf{r}_0) $ into a tractable prior standard gaussian distribution  $q\left(\mathbf{r}_T\right) \sim \mathcal{N}\left(\mathbf{r}_T; \mathbf{0}, \mathbf{I}\right)$.
In the forward process, there are no learnable parameters and we can obtain every intermediate state $\textbf{r}_t$ by just adding a quantity of noise defined by the noise scheduler to $\textbf{r}_0$:
\begin{equation}
q\left(\mathbf{r}_t \mid \mathbf{r}_0\right):=\mathcal{N}\left(\mathbf{r}_t ; \sqrt{\bar{\alpha}_t} \mathbf{r}_0,\left(1-\bar{\alpha}_t\right) \mathbf{I}\right) 
\label{2}
\end{equation}
\begin{equation}
    \bar{\alpha_t}:=\prod_{s=0}^t \alpha_s
\end{equation}
Using the reparametrization trick, eq.~\eqref{2}, can be written as follow:
\begin{equation}
    \mathbf{r}_t = \sqrt{\bar{\alpha_t}}\mathbf{r}_0 + \sqrt{1-\bar{\alpha_t}} \bm{\epsilon} \qquad \bm{\epsilon} \sim \mathcal{N}(\mathbf{0},\mathbf{I})
    \label{reparametrization}
\end{equation}
The inverse procedure models another Markov chain that starting from pure noise $\mathbf{r}_T$ and conditioned to $\mathbf{x}_I$, progressively eliminates noise in a recursive manner to predict the exact residual image $\mathbf{r}_0$. This iterative noise elimination is referred to as the conditional reverse transition $p_\theta(\textbf{r}_{t-1}|\textbf{r}_t,\textbf{x}_I)$, which is estimated by optimizing the trainable parameters $\theta$ within a denoising network:
\begin{equation}
    p_\theta\left(\mathbf{r}_{t-1} \mid \mathbf{r}_t,\textbf{x}_I\right):=\mathcal{N}\left(\mathbf{r}_{t-1} ; \bm{\mu}_\theta\left(\mathbf{r}_t, t,\textbf{x}_I\right), \bm{\Sigma}_\theta\left(\mathbf{r}_t, t,\textbf{x}_I\right)\right)
    \label{back_kernel}
\end{equation}
As denoising network $f_\theta$, we design a novel and effective variation of NAFNet that, along with $\textbf{r}_t$ and $\textbf{x}_I$ as inputs, takes into consideration also the conditioning of the timestep $t$ of diffusion models in order to improve performance in denoising and deblurring tasks. Going into details, following \cite{Refusion}, we employ an additional MLP to model the time embedding and transform it into shift parameters $\gamma$ and $\beta$. These parameters control scale and bias terms of each normalization layer that now are conditioned on $t$. The mechanism can be easily visualized in Fig. \ref{modifiedNafblock}.

Given a noisy residual sample $\textbf{r}_t$, we train the denoiser network $f_\theta$ to directly estimate $\textbf{r}_0$ instead of estimating the amount of noise needed to form $\textbf{r}_{t-1}$. The prediction of  $\textbf{r}_0$ and $\textbf{r}_{t-1}$ is equivalent as they can be transformed into each other
through eq.~\eqref{reparametrization}.
It has been proven that this approach reduces diversity in favor of generation quality in the first denoising steps \cite{palette,sr3 ,DocDiff}. 
We associate to this component two dedicated loss functions. The first one that tries to minimize the distance between the estimated reverse transition $p_\theta(\textbf{r}_{t-1} \mid \textbf{r}_t,\textbf{x}_I)$ and the real one $q(\textbf{r}_{t-1} \mid \textbf{r}_t,\textbf{x}_I)$. The second one tries to specialize the denoiser network to recover high-frequency details.
\vspace{0.1cm}
\begin{equation}
    L_{\text{DPM}} = \mathbb{E}\left\|\textbf{r}_0-f_\theta\left(\textbf{r}_t, t, \textbf{x}_I\right)\right\|_2
\end{equation}
\begin{equation}
    L_{\text{HIGH}} =   \mathbb{E}\left\|\phi_H \ast\left(\textbf{r}_0-f_\theta\left(\textbf{r}_t, t, \textbf{x}_I\right)\right)\right\|_2
\end{equation}
\begin{equation}
    L_{\text{DENOISER}} = L_{\text{DPM}} + 2 \ast L_{\text{HIGH}}
    \label{lossdenoiser}
\end{equation}
Overall, the used loss function is the combination of eq.~\eqref{lossinit} and eq.~\eqref{lossdenoiser}:
\begin{equation}
    L_{\text{TOT}} = \frac{1}{2}L_{\text{INIT}} + L_{\text{DENOISER}}
    \label{ltot}
    \vspace{0.1cm}
\end{equation}
The training strategy adopted is detailed described using pseudo-code in Algorithm~\ref{alg:Training}.
\begin{algorithm}[t]
\caption{Training NAF-DPM}\label{alg:Training}
\begin{algorithmic}
\vspace{0.1cm}
\STATE \textbf{INPUT}: $i_\theta$ initial predictor, $f_\theta$ denoiser network, $T$ \hspace{0.01cm} Sampling steps, \hspace{0.01cm} $\bar{\bm{\alpha}}_{0:T}$ noise scheduler;
\vspace{0.1cm}
\STATE \textbf{OUTPUT}:  $i_{\theta^\star}$ trained initial predictor, $f_{\theta^\star}$ trained denoiser network;
\vspace{0.3cm}
\STATE \textbf{FOR} $i \gets 0$\textbf{ TO} MAX\_ITERATIONS\textbf{ DO}:
\STATE \hspace{0.5cm} ($\textbf{x}_{gt}$,\textbf{y}) $\sim$ q(data) \hspace{1cm}  sample images from dataset
\vspace{0.1cm}
\STATE \hspace{0.5cm} $t$ $\sim$ Uniform({1,...,T}) \hspace{1.87cm}  sample timestep
\vspace{0.1cm}
\STATE \hspace{0.5cm} $\bm{\epsilon}  \sim \mathcal{N}(\textbf{0},\textbf{I})$ \hspace{3.08cm}  sample noise
\vspace{0.1cm}
\STATE \hspace{0.5cm} $\textbf{x}_I \gets i_\theta(\textbf{y})$ \hspace{3.13cm}  initial prediction
\vspace{0.1cm}
\STATE \hspace{0.5cm} $\textbf{r}_t = \sqrt{\bar{\alpha_t}} (\textbf{x}_{gt} - \textbf{x}_I) + \sqrt{1- \bar{\alpha_t}}\bm{\epsilon}$ \hspace{0.06cm} forward diffusion
\vspace{0.1cm}
\STATE \hspace{0.5cm} $\tilde{\textbf{r}_0 }= f_\theta(\textbf{r}_t,t,\textbf{x}_I)$ \hspace{2.47cm}residual estimation\vspace{0.1cm}
\STATE \hspace{0.5cm} $L_{\text{TOT}}(\textbf{x}_{gt},\textbf{x}_I,\tilde{\textbf{r}_0})$ \hspace{2.40cm} loss computation
\vspace{0.1cm}
\STATE \hspace{0.5cm} Take a gradient descend update on $i_\theta,f_\theta$ using $L_{\text{TOT}}$
\vspace{0.1cm}
\STATE \textbf{RETURN} $I_{\theta^\star}$, $f_{\theta^\star}$
\end{algorithmic}
\label{alg2}
\end{algorithm}

\subsection{Fast Sampling Strategy}
Diffusion models are very powerful generative models but their main weakness lies in their generative slowness since sampling from them requires hundreds or even thousands of neural network evaluations. One of the first works that tries to speed up the generative process of DPM is the strategy introduced by Song \textit{et al}. in \cite{DDIM}.  They introduced a non-Markovian definition of diffusion models with the same DPM training objective, but whose reverse process can be much faster to sample from. Surprisingly, their reverse kernel definition also allows for a deterministic definition, called denoising diffusion implicit model (DDIM):
\begin{equation}
    \begin{array}{r}
\mathbf{r}_{t-1}=\sqrt{\bar{\alpha}_{t-1}}\tilde{\textbf{r}_0 }
+\sqrt{1-\bar{\alpha}_{t-1}} \frac{\textbf{r}_t-\sqrt{\bar{\alpha}_{t}} \tilde{\textbf{r}_0}}{\sqrt{1-\bar{\alpha}_{t}}}
\end{array}
\label{DDIM}
\end{equation}
\begin{equation}
     \tilde{\textbf{r}}_0 = f_\theta\left(\textbf{r}_t,t,\textbf{x}_I\right)
\end{equation}
This is one of the first works that successfully produce higher-quality images with fewer denoising steps compared to DPM.
Taking this approach as a baseline,
in our work, we adopt a different method to speed up the sampling phase without sacrificing performance: we employ a fast high-order solver for diffusion ODEs called DPM-solver \cite{dpmsolver}. It has a convergence order guarantee in 10-20 function evaluations. We decided to use this type of solver since document enhancement does not require a high degree of diversity and modelling the sampling phase from our DPM as solving the corresponding diffusion ordinary differential equation (ODE) could be a suitable choice.
The formulation of DPM-solver assumes a continue-time DPM but we defined and trained a discrete-time DPM. However, the authors described formally how to switch from one configuration to another and give black-box coding support for this operation.  The overall sampling strategy is defined in Alg.~\ref{alg:Sampling}.

\begin{algorithm}[t]
\caption{Sampling from NAF-DPM}\label{alg:Sampling}
\begin{algorithmic}
\STATE \textbf{INPUT}: $i_{\theta^\star}$ initial predictor, $f_{\theta^\star}$ denoiser network, $T$ \hspace{0.01cm} sampling steps, \hspace{0.01cm} $\bar{\bm{\alpha}}_{0:T}$ noise scheduler, $\textbf{y}$ \hspace{0.01cm} degraded image;
\STATE \textbf{OUTPUT}: restored images generated by NAF-DPM;
\vspace{0.2cm}
\STATE \textbf{SAMPLING WITH DDIM:}
\vspace{0.2cm}
\STATE \hspace{0.5cm}$\textbf{x}_I \gets i_\theta(\textbf{y})$ \hspace{0.3cm} first prediction from initial predictor
\STATE \hspace{0.5cm}$\textbf{r}_T \sim \mathcal{N}(\textbf{0},\textbf{I}) $ \hspace{0.3cm} sample from normal distribution
\STATE \hspace{0.5cm}\textbf{FOR} $t \gets T$\textbf{ TO} $1$\textbf{ DO}:
\vspace{0.2cm}
\STATE \hspace{1cm} $ \tilde{\textbf{r}}_0 \gets f_\theta\left(\textbf{r}_t,t,\textbf{x}_I\right) $ \hspace{1.5cm} denoiser estimation
\vspace{0.2cm}
\STATE \hspace{1cm} $ \bm{\epsilon}_t \gets \frac{\textbf{r}_t-\sqrt{\bar{\alpha}_{t}} \tilde{\textbf{r}_0}}{\sqrt{1-\bar{\alpha}_{t}}}$ \hspace{2.25cm} noise computation
\vspace{0.2cm}
\STATE \hspace{1cm} $\textbf{r}_{t-1}=\sqrt{\bar{\alpha}_{t-1}}\tilde{\textbf{r}}_0 
+\sqrt{1-\bar{\alpha}_{t-1}} \bm{\epsilon}_t \hspace{0.2cm}$ refinement
\vspace{0.2cm}
\STATE\hspace{0.5cm}\textbf{END FOR}
\STATE\hspace{0.5cm}\textbf{RETURN}\hspace{0.20cm}$\textbf{x}_I + \textbf{r}_0$ \hspace{0.1cm} initial prediction + residual image

\vspace{0.2cm}
\STATE \textbf{SAMPLING WITH DPM-solver:}
\vspace{0.2cm}
\STATE \hspace{0.5cm}$\textbf{x}_I \gets i_\theta(\textbf{y})$ \hspace{0.3cm} first prediction from initial predictor
\vspace{0.2cm}
\STATE \hspace{0.5cm}$\textbf{r}_T \sim \mathcal{N}(\textbf{0},\textbf{I}) $ \hspace{0.3cm} sample from normal distribution
\vspace{0.2cm}
\STATE \hspace{0.5cm}$\textbf{r}_0 \gets$ DPM-solver( $\bar{\bm{\alpha}}_{0:T}$,  $f_\theta$(.), $\textbf{r}_T$, $T$, $\textbf{x}_I$) \hspace{0.3cm} Residual 
\vspace{0.2cm}
\STATE\hspace{0.5cm}\textbf{RETURN}\hspace{0.20cm}$\textbf{x}_I + \textbf{r}_0$ \hspace{0.1cm} initial prediction + residual image

\end{algorithmic}
\label{alg1}
\end{algorithm}

\subsection{Differentiable OCR-Guided Finetuning}
\label{CRNN}
The great generative power of the denoiser network is capable of recovering high-frequency details from images semi-reconstructed from the initial predictor. In the case of text documents, when sometimes source characters are poorly defined, such power can lead to recovering wrong characters. In fact, during experiments, we noticed that some characters were wrong, although well reconstructed and well recognized by an OCR system. To limit this phenomenon we introduce an additional loss term that tries to push the entire network to reduce the errors in the characters. The main idea is to use a commercial OCR system. Fed this OCR with the enhanced images from the network and computed the Levenshtein distance between the extracted text and reference text. In the end, propagate back the error to make the network aware of mistakes made on characters. There are two problems with this approach: at first the Levenshtein distance is not differentiable and so it can not be used as a loss function.
\begin{figure}[!ht]
\centering
\includegraphics[width=\linewidth]{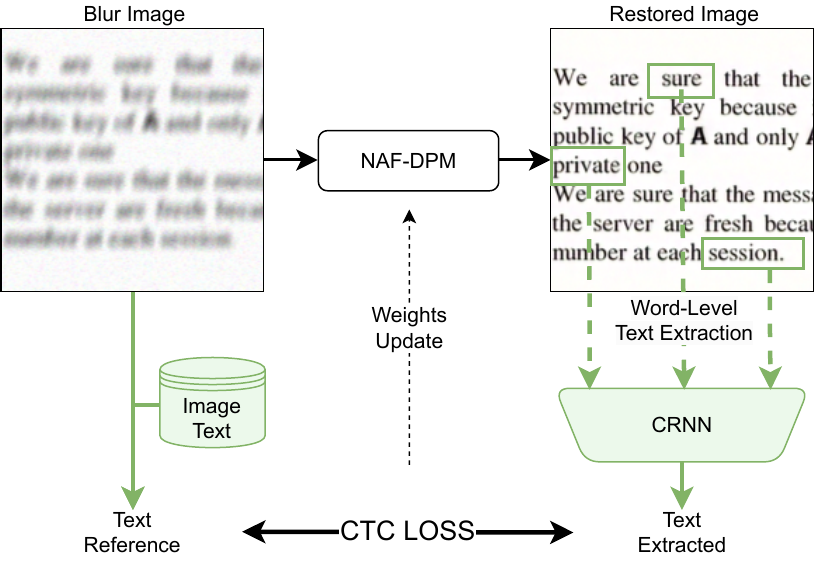}
\caption{Our proposed OCR-based finetuning enhances the character readability and reduces the character error rate made by our network during the sampling phase. We attach a CRNN module just after the output layer of our network. We perform word-level text extraction and we feed the CRNN module with all the extracted word-level image patches. The CRNN emulates the behaviour of a commercial OCR and recognizes text from input patches. The text is compared to the reference one and CTC Loss Function is computed. In the end, the weights of our network are updated using eq.~\eqref{loss_finetune}.}
\label{architecture_CRNN}
\end{figure}
Then, commercial OCR systems could contain non-differentiable modules that block the gradient backpropagation. To solve the first problem we employ connectionist temporal classification (CTC) loss function. CTC is a machine learning algorithm commonly used for sequence-to-sequence tasks. Given an input sequence $\textbf{A}$ and an output sequence $\textbf{B}$, the goal of the CTC is to learn a mapping function $P(\textbf{\textbf{B}}|\textbf{A})$ that maximizes the probability of the correct output sequence given the input sequence. 
To solve the second problem we replace a commercial OCR system with a convolutional recurrent neural network (CRNN \cite{CRNN}) that is pretrained to emulate the behaviour of the OCR, following the work by Randika \textit{et al}. \cite{unknown-Box}.

As illustrated in Fig.~\ref{architecture_CRNN}, we place the CRNN just after the network. From the restored image we extract word-level text that is fed into CRNN. The output of this module is compared to reference text and CTC loss is computed. 
Since this process requires additional training time with respect to traditional training we decided to enable the CRNN components only after a predefined number of iterations of standard training; a sort of finetuning strategy after having trained the main network for a predefined number of iterations.
\begin{equation}
    L_{\text{CTC}} = \text{CTC(text}_{\text{extracted}},\text{text}_{\text{reference}})
\end{equation}
\begin{equation}
    L_{\text{FINETUNE}} = L_{\text{TOT}} + L_{\text{CTC}}
    \label{loss_finetune}
    \vspace{0.1cm}
\end{equation}

%
%

\section{Experiments and Results}

\subsection{Document Deblurring}
We begin our experiments with document deblurring. For this task, the blurry document images text dataset is used \cite{DeblurringDataset}. It is composed of 66,000 blurry text images of size 300x300 with their corresponding ground truth counterpart. It includes also another section composed of 94 images of size 512x512 used for OCR testing. We randomly select 30,000 images for training and 10,000 images for validating our model.
During training, we randomly extract from each image patches of size 128x128 performing also random horizontal flip and random rotation. We jointly train the initial predictor network and denoiser network for 800k iterations trying to reduce the loss function eq.~\eqref{ltot}. We fix the timesteps hyperparameter to $T=100$. During sampling we use both DDIM sampling algorithm, eq.~\eqref{DDIM}, and DPM-solver, to compare the results coming out from the two approaches. To assess network performance we use both pixel-level metrics (PSNR and SSIM \cite{PSNR-SSIM}), and perceptual similarity metrics (LPIPS \cite{LPIPS} and DISTS \cite{DISTS}).

\begin{table}[h]
\centering
\caption{Ablation studies on sampling methods on the Document Deblurring Dataset Test Dataset (10k randomly selected 300x300 images) \cite{DeblurringDataset}.}
\resizebox{\columnwidth}{!}{%
\begin{tabular}{ccccccc}
\toprule
\begin{tabular}[c]{@{}c@{}}Sampling\\ Method\end{tabular} & \begin{tabular}[c]{@{}c@{}}Sampling \\ Iterations\end{tabular} & PSNR ($\uparrow$)  & SSIM ($\uparrow$) & LPIPS ($\downarrow$) &  DISTS ($\downarrow$)  \\ \midrule
DDIM              & 100 & 25.819 & 0.964& 0.0235  &  0.0497      \\     
DPM-Solver        & 100 & 27.405& 0.974 & 0.0211 & 0.0406 &                              \\
DPM-Solver        & 50  & 27.511 & 0.975 & 0.0207 & 0.0400       \\
DPM-Solver        & 20  & 27.588 & 0.975 & \textbf{0.0204} & \textbf{0.0398}       \\
DPM-Solver        & 10  & \textbf{27.607} & \textbf{0.975} & 0.0207 & 0.0409       \\ \bottomrule
\end{tabular}%
}
\vspace{0.1cm}

\label{tab:ablation study}
\end{table}

From the results shown in the Table~\ref{tab:ablation study}, we can highlight the superiority of the sampling method introduced by \cite{dpmsolver}. This method is able to produce images of superior quality from the point of view of all metrics examined. In particular, the PSNR metric is much higher using DPM-solver, a sign that the predicted images are more accurate with respect to the generated ones using the DDIM sampling strategy.
As warranted by the authors of \cite{dpmsolver}, the method converges in 10-20 sampling steps. As the number of steps increases, there is a slight decrease in pixel-level metrics while perceptual ones remain stable. 

\begin{figure*}[!t]
\centering
\subfloat{\includegraphics[width=2\columnwidth]{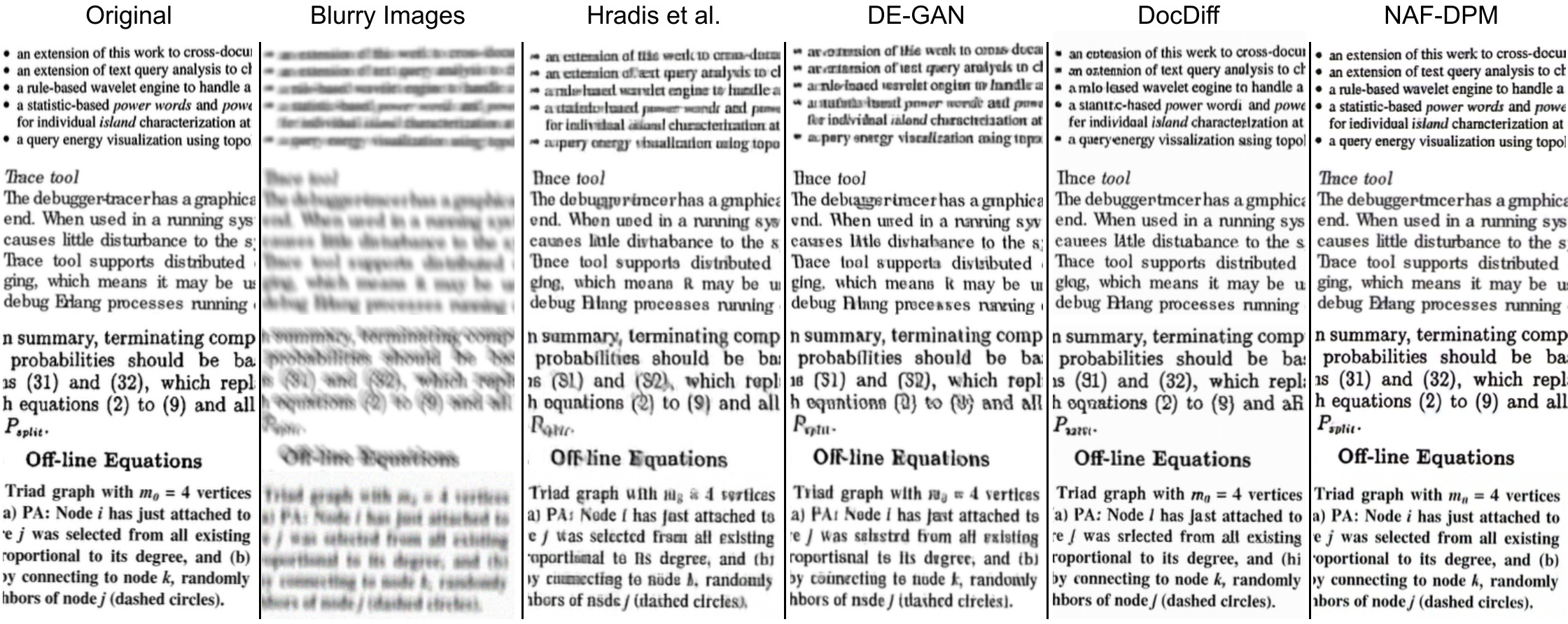}}%
\label{fig_first_case}
\caption{Document Deblurring: Comparison between our method and other approaches. From left to right it can be seen the blurred images, and the restored images by Hradis \textit{et al}. \cite{DeblurringDataset}, DE-GAN\cite{DEGAN}, DOC-DIFF \cite{DocDiff}, our network and in the last column the original images. The characters restored by our network are clearly more readable and accurate. The Generative Adversarial Network approach produces confused and blurred characters that bring poor performances from the OCR point of view. DOC-DIFF based on the approach "Predict and refine" is able to produce more accurate images. The characters are not so blurred as in DEGAN, however, some of them are substituted by the wrong character. This behaviour can be due to the high diversity property of a diffusion model. In our approach, a better suitable backbone network and an improved training-finetuned strategy effectively reduce the occurrence of substituted characters with a clear advantage for the accuracy of an OCR system.  }
\label{fig_sim}
\end{figure*}

Nevertheless, we will compare our approach with results achieved by state-of-the-art methods in the document deblurring problem. For this type of analysis, we use the document deblurring OCR test dataset that, as previously mentioned, contains 94 images of size 500x500. We sample using a DPM-solver set at 20 sampling iterations. Results are listed in Table~\ref{tab:comparison other methods}.
NAF-DPM beats all pre-existing methods based on convolutional neural network (Hradis \textit{et al} \cite{DeblurringDataset}), conditional generative adversarial network (DE-GAN \cite{DEGAN}) and diffusion probabilistic model (DocDiff \cite{DocDiff}) by a large margin. In particular, in comparison with DocDiff, it can be seen a gain of more than 4db in PSNR metrics and much more important, there is a clear halving on perceptual similarity metrics (LPIPS and DISTS).
\begin{table}[t]
\centering
\caption{Results of image deblurring on OCR TEST Dataset (94 images of size 500x500) \cite{DeblurringDataset}.}
\resizebox{\columnwidth}{!}{%
\begin{tabular}{ccccccc}
\midrule
Method  & Model &PSNR ($\uparrow$)  & SSIM ($\uparrow$) & LPIPS ($\downarrow$) &  DISTS ($\downarrow$) \\ \midrule
Hradis \cite{DeblurringDataset} &CNN  & 30.629 & 0.987 & 0.0135 & 0.0258  \\     
DE-GAN \cite{DEGAN}      &cGAN        & 28.803 & 0.985 & 0.0144 & 0.0237       \\
DocDiff \cite{DocDiff}   &DPM          & 29.787 & 0.989 & 0.0094 & 0.0339        \\
\textbf{NAF-DPM}  &DPM         & \textbf{34.377} & \textbf{0.994} & \textbf{0.0046} & \textbf{0.0228}    \\ \bottomrule
\end{tabular}%
}

\label{tab:comparison other methods}
\vspace{-0.2cm}
\end{table}
Visual results can be appreciated from Fig.~\ref{fig_sim}. It can be seen that images produced by pure CNN and GAN networks have poorly defined text with blurred and distorted edges (Hradis \textit{et al}. \cite{DeblurringDataset} and DE-GAN \cite{DEGAN}). The strategy ``predict and refine" adopted by DocDiff \cite{DocDiff} and by our network allows more readable text to be retrieved. Unlike DocDiff, our approach seems to be much more effective in retrieving all the details and text edges. Credit can be given to the choice of backbone networks. Using a designated and specialized network for deburring and denoising tasks allows for more precise initial prediction and consequently also for more specific refinement by the diffusion model.  In addition, we have also a gain in the sampling time, using DPM-solver we are able to retrieve the vast majority of details in just 10-20 function evaluations (in Fig.~\ref{fig_sim} images from our network sampled in 20 DPM-solver sampling iterations) with respect to DocDiff that requires more than 50 function evaluation to make text edges smoother and sharper (in Fig.~\ref{fig_sim} images from DocDiff sampled in 100 DDIM stochastic sampling iterations).

\subsection{OCR Evaluation}
In this section, we focus our attention on the performance of OCR on degraded and enhanced documents. We compare the results from NAF-DPM with state-of-the-art results in the document deblurring OCR test dataset \cite{DeblurringDataset} using PaddleOCR\footnote{https://github.com/PaddlePaddle/PaddleOCR} as OCR. We have manually labelled the dataset since the original labelling present in \cite{DeblurringDataset} was done using an old OCR system on clean images and indeed it presents a lot of errors and missing characters. The results are shown in Table~\ref{tab:results cer and psnr}.
Our network is able to substantially reduce the error on characters made by PaddleOcr, irrefutable evidence that images are well restored and high-frequency details, fundamentals for the right detection of characters, are preserved. We almost halve the character error rate (CER) compared to DocDiff \cite{DocDiff}. Compared to the other two benchmarks we have a performance increase of 3.88x with respect to Hradis \textit{et al}. \cite{DeblurringDataset} and 4.90x with respect to DE-GAN \cite{DEGAN}. These results define a new state-of-the-art (SOTA) for document deblurring OCR test dataset.

In order to further reduce the CER metric, we finetune our network using an additional CRNN network and the augmented loss function, eq.~\ref{loss_finetune}. The adopted dataset is the Document Deblurring dataset \cite{DeblurringDataset}. From the patches we extract text and corresponding bounding boxes using Paddle-OCR. The extraction is done offline since it requires a non-negligible amount of time.
We pretrain a CRNN Network on the training patches and the correspondent extracted text for 20 epochs. In this way, the behaviour of PaddleOCR is emulated by a differentiable CRNN module \cite{CRNN}.
We take model checkpoint at 700k iterations and we finetune it for an additional 100k, so that at the end we can compare the performance of the two models at equal training state (i.e., 800k training steps).

\begin{figure*}[!ht]
\centering
\includegraphics[width=1.9\columnwidth]{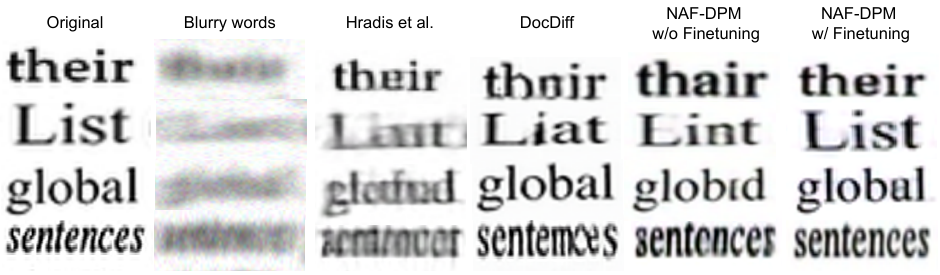}
\caption{Comparison between our method and other approaches on manually selected words from images of document deblurring OCR test dataset \cite{DeblurringDataset}}.
\label{Words comparison}
\end{figure*}

\begin{table}[t]
\centering
\caption{Character error rate comparison among different methods on document deblurring OCR test dataset \cite{DeblurringDataset}}
\resizebox{0.6\columnwidth}{!}{%
\begin{tabular}{ccc}
\midrule
Method  & PSNR ($\uparrow$)   & CER ($\downarrow$)  \\ \midrule
Hradis \cite{DeblurringDataset}   & 30.629 & 5.44  \\     
DE-GAN \cite{DEGAN}              & 28.803 &    6.87    \\
DocDiff \cite{DocDiff}             & 29.787 & 2.78      \\ 
\textbf{NAF-DPM}          & \textbf{34.377} &  \textbf{1.55} \\ 
\midrule
\end{tabular}%
}
\label{tab:results cer and psnr}
\vspace{-0.3cm}
\end{table}
The addition of the CRNN module and the new loss function would seem to bring marginal improvements in the CER metric (see Table~\ref{tab:results cer}). In particular, the results show that this approach succeeds in better training the network and making it more aware of the characters it is reconstructing. From both the decreasing CER metric and the qualitative results shown in Fig.~\ref{Words comparison} we note that there are some characters exactly reconstructed only in the finetuned version of NAF-DPM.
\begin{table}[ht]
\centering
\caption{Character error rate comparison between different methods on document deblurring OCR test dataset \cite{DeblurringDataset}}
\resizebox{0.7\columnwidth}{!}{%
\begin{tabular}{cc}
\midrule
Method    & CER ($\downarrow$)  \\ \midrule
NAF-DPM w/o finetuning         &  1.55 \\ 
NAF-DPM w/ finetuning & \textbf{1.40}  \\
\midrule
\end{tabular}%
}
\label{tab:results cer}
\end{table}

\subsection{Binarization}
We continue our experiments by investigating the robustness of our method to a different task: document binarization. We train and evaluate our network on three datasets from the annual Document Image Binarization Competition (DIBCO). In particular we choose DIBCO2017 \cite{DIBCO2017}, H-DIBCO2018 \cite{DIBCO2018} and the challenging DIBCO2019 \cite{DIBCO2019}.  These datasets contain
degraded document images with handwritten and printed text. DIBCO 2019 contains also ancient documents written on papyrus, which makes this dataset highly challenging.
In each experiment, we pick a dataset as the test dataset (e.g. DIBCO2017) and we construct the training dataset with the remaining data from annual DIBCOs (from DIBCO2009 to DIBCO2019). We integrate the training material with the Bickley Diary dataset \cite{birkley}, Persian Heritage Image Binarization Dataset (PHIDB) \cite{persian}, the Synchromedia Multispectral dataset (S-MS) \cite{synch} and Palm Leaf dataset \cite{palm}. From each image of the training set we extract 256x256 overlapped patches following the work in \cite{docentr} and for each experiment, we jointly train the initial predictor and denoiser for 100000 iterations. We sample final images using DPM-solver set at 11 sampling iterations.
\subsubsection{Quantitative Evaluation}
Standard metrics used to evaluate the performance of the model are: Peak signal-to-noise ratio (PSNR), F-measure and pseudo-F-measure ($F_{ps}$).

Table~\ref{tab:dibco20172018} shows the comparison results of different methods on DIBCO 2017 test set \cite{DIBCO2017}. Out of these results, we can say that our method is superior to the current state-of-the-art methods according to PSNR metric. DE-GAN \cite{DEGAN} keeps the best performance in F-measure and $F_{ps}$, although the official results presented by the authors would seem to outperform the trend of various models over the years by a large margin. 
Despite this, NAF-DPM establishes itself as the best among the others, demonstrating great robustness and stability on the various degradations present in these documents.  
\begin{table*}[t]
\centering
\caption{Results of image binarization on DIBCO 2017 and H-DIBCO 2018 Databases \cite{DIBCO2017, DIBCO2018}. NAF-DPM achieves excellent and consistent results in the various examined datasets. The best algorithms of competitions \cite{DIBCO2017, DIBCO2018} use pre-processing and post-processing steps to achieve good performance. Jemni \textit{et al} use additional pertaining data to specialize their network to handwritten document enhancement. In bold the best results, underlined the second best.
 }
\begin{tabular}{lcccccccccc}
\toprule
\multicolumn{1}{c}{\multirow{2}{*}{Method}} & 
\multicolumn{1}{c}{\multirow{2}{*}{Model}} & 
\multicolumn{1}{c}{\multirow{2}{*}{Parameters}} &
\multicolumn{1}{c}{\multirow{2}{*}{\parbox{1.5cm}{Additional Processing}}}&
\multicolumn{3}{c}{DIBCO 2017}&\multicolumn{3}{c}{DIBCO 2018} \\ 
\cmidrule(l){5-7} 
\cmidrule(l){8-10} 
\multicolumn{1}{c}{}  &
\multicolumn{1}{c}{}  &
\multicolumn{1}{c}{} 
& \multicolumn{1}{c}{}                       
& PSNR($\uparrow$)   
& F-Measure($\uparrow$)   
& F$_{ps}$($\uparrow$)                                            
& PSNR($\uparrow$)   
& F-Measure($\uparrow$)   
& F$_{ps}$($\uparrow$)\\ \midrule
Competition Top    \cite{DIBCO2017,DIBCO2018}               &CNN          &-  &\cmark   &  18.28  & 91.04      & 92.86   &  19.11  & 88.34  & 90.24\\    
Jemni \textit{et al}. \cite{enhancetoread} &cGAN &15.8M & \cmark   & 17.45 & 89.80 & 89.95 &20.18& 92.41 & 94.35\\
\midrule
Otsu        \cite{otsu}           &Thresh     &-  &\xmark   &   13.85 &  77.73  & 77.89  &   9.74 &  51.45  & 53.05    \\
Sauvola     \cite{SAUVOLA}        &Thresh       &-  &\xmark   &   14.25 &  77.11  & 84.10   & 13.78&67.81&74.08  \\
 
DE-GAN \cite{DEGAN}               &cGAN         &31M  &\xmark  &  18.74  &  $\textbf{97.91}^\ast$ & \textbf{98.23}$^\ast$   &  16.16 &  77.59    & 85.74        \\     

DocEnTr \cite{docentr}            &Vi-Trans     &68M  &\xmark   &  19.11      &      92.53       &   95.15  &19.46&90.59&93.97     \\
D$^2$BFormer \cite{dbformer}      &Vi-Trans     &194M  &\xmark  &19.35&93.52&95.09&18.91&88.84&93.42 \\ 
DocDiff \cite{DocDiff}           &DPM          &8.2M  &\xmark &-&-&-&17.92&88.11&90.43\\  
\textbf{NAF-DPM}                  &DPM          &9.4M  &\xmark   & \textbf{19.40} & \underline{93.55}      & \underline{95.76}    &\textbf{19.67}&\textbf{90.64}&\textbf{94.51}    \\ \bottomrule
\end{tabular}
\label{tab:dibco20172018}
\end{table*}

We continue the quantitative analysis of our model on the most recent H-DIBCO 2018 \cite{DIBCO2018} and DIBCO 2019 \cite{DIBCO2019} datasets. 
In H-DIBCO 2018 \cite{DIBCO2018} our network achieves state-of-the-art performance in terms of pseudo-F-Measure, while the model introduced by Jemini \textit{et al}. \cite{enhancetoread} keeps the best results in PSNR and F-Measure. It is important to note that their model is designed to restore handwritten documents and uses additional training data. Despite this, NAF-DPM can reach similar results without any additional training data, pre-processing, post-processing stages, or fine-tuning strategies. Results are shown in Table~\ref{tab:dibco20172018}.

Regarding the challenging DIBCO 2019 \cite{DIBCO2019}, our approach clearly reaches top performance in all the metrics, beating both general purpose binarization algorithm \cite{otsu,SAUVOLA} and specifically designed algorithm for this kind of dataset \cite{DIBCO2019,dbformer}. Results are shown in Table~\ref{tab:dibco2019}.

This study leads us to say that our method turns out to be very robust to different forms of degradation. In each dataset, it excels in at least one metric and performs as the first or second best network each time, unlike other networks, such as DEGAN\cite{DEGAN}, which turn out to be SOTA for one dataset (DEGAN-DIBCO2017) but perform worse in other datasets (H-DIBCO2018 and DIBCO2019).
The ``predict and refine" strategy is also in this case very effective and shows clearly the superiority with respect to common and well-known approaches such as generative adversarial networks (GANs) and vision transformers.

It is interesting to note the increase in performance achieved by NAF-DPM against DocDiff \cite{DocDiff}, which is also based on DPM. The design of dedicated nonlinear activation-free networks for the initial predictor and the denoiser along with the use of an ODE fast solver allows to unlock the full potential of DPMs for these types of tasks.
This is confirmed by the clear margin in the PSNR metrics in both the H-DIBCO 2018 \cite{DIBCO2018} dataset (from 17.92db to 19.60db) and in DIBCO 2019 \cite{DIBCO2019}(from 15.14db to 15.40db).

\begin{table}[t]
\centering
\caption{Results of image binarization on DIBCO 2019 Database \cite{DIBCO2019}.  NAF-DPM outperforms all other methods in all metrics, demonstrating a great power of generalization. In bold the best results, underlined the second best.}
\begin{tabular}{lcccc}
\toprule
\multicolumn{1}{c}{\multirow{2}{*}{Method}} & \multicolumn{1}{c}{\multirow{2}{*}{Model}} & \multicolumn{3}{c}{DIBCO 2019} \\ \cmidrule(l){3-5} 
\multicolumn{1}{c}{}                        & \multicolumn{1}{c}{} 
                   & PSNR($\uparrow$)  & F-Measure($\uparrow$)  & F$_{ps}$($\uparrow$) \\ \midrule
Otsu     \cite{otsu}        & Thresh        &   9.08  &  47.83  & 45.59        \\
Sauvola     \cite{SAUVOLA}   & Thresh       &  13.72  &  51.73  & 55.15        \\
Competition Top  &CNN  &  14.48  & 72.88   & 72.15        \\
DE-GAN \cite{DEGAN}  &cGAN               &  12.29  &  55.98  & 53.44        \\
D$^2$BFormer \cite{dbformer} &Vi-Trans&15.05&67.63&66.69 \\
DocDiff    \cite{DocDiff} &DPM                        &   \underline{15.14} &  \underline{73.38}  & \underline{75.12}        \\ 
           \textbf{NAF-DPM}        &DPM             & \textbf{15.39} & \textbf{74.61}     &\textbf{ 76.25  }      \\  \bottomrule
\end{tabular}
\vspace{0.1cm}
\label{tab:dibco2019}
\end{table}

To strengthen the power of our approach, we can analyze the results obtained in relation to the number of parameters of the various networks. In Table~\ref{tab:dibco20172018}, 
we can appreciate that the performance of networks is not correlated with the increasing number of their parameters. Our approach, together with DocDiff \cite{DocDiff}, managed to achieve very good results while keeping the number of parameters very small. Our network is much more powerful than DE-GAN \cite{DEGAN} by having one-third of the parameters. It manages to exceed the performance of the best-known vision transformer approaches dedicated to this task having about one-sixth of the parameters of DocEnTr \cite{docentr} and even one-twentieth compared to D$^2$BFormer \cite{dbformer}.
Acronyms used in Tables ~\ref{tab:dibco20172018},~\ref{tab:dibco2019}: Thresh: Thresholding, CNN: Convolutional Neural Network, cGAN: Conditioned Generative Adversarial Network, Vi-Trans: Vision Transformer, DPM: Diffusion Probabilistic Model.

\subsubsection{Qualitative Evaluation}
\begin{figure}[t]
\centering
\includegraphics[width= \linewidth]{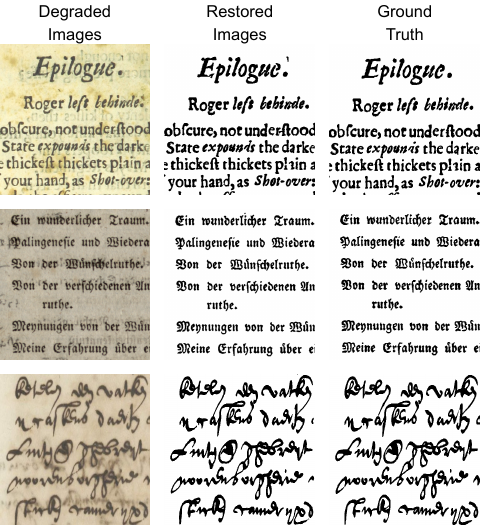}
\caption{Visual results. Three random extracted patches from DIBCO 2017 \cite{DIBCO2017}. From left to right we find Degraded images, Restored images, and ground truth images. Reconstructed images are almost identical to ground truth.}
\label{dibco2017 visual results}
\end{figure}

In Figure~\ref{dibco2017 visual results} there are visual results from 3 different patches of DIBCO2017. Binarized images are very similar to ground truth. This confirms the high values of PSNR and F-Measure.

\begin{figure}[h]

\centering
\includegraphics[width=\linewidth,height=19cm]{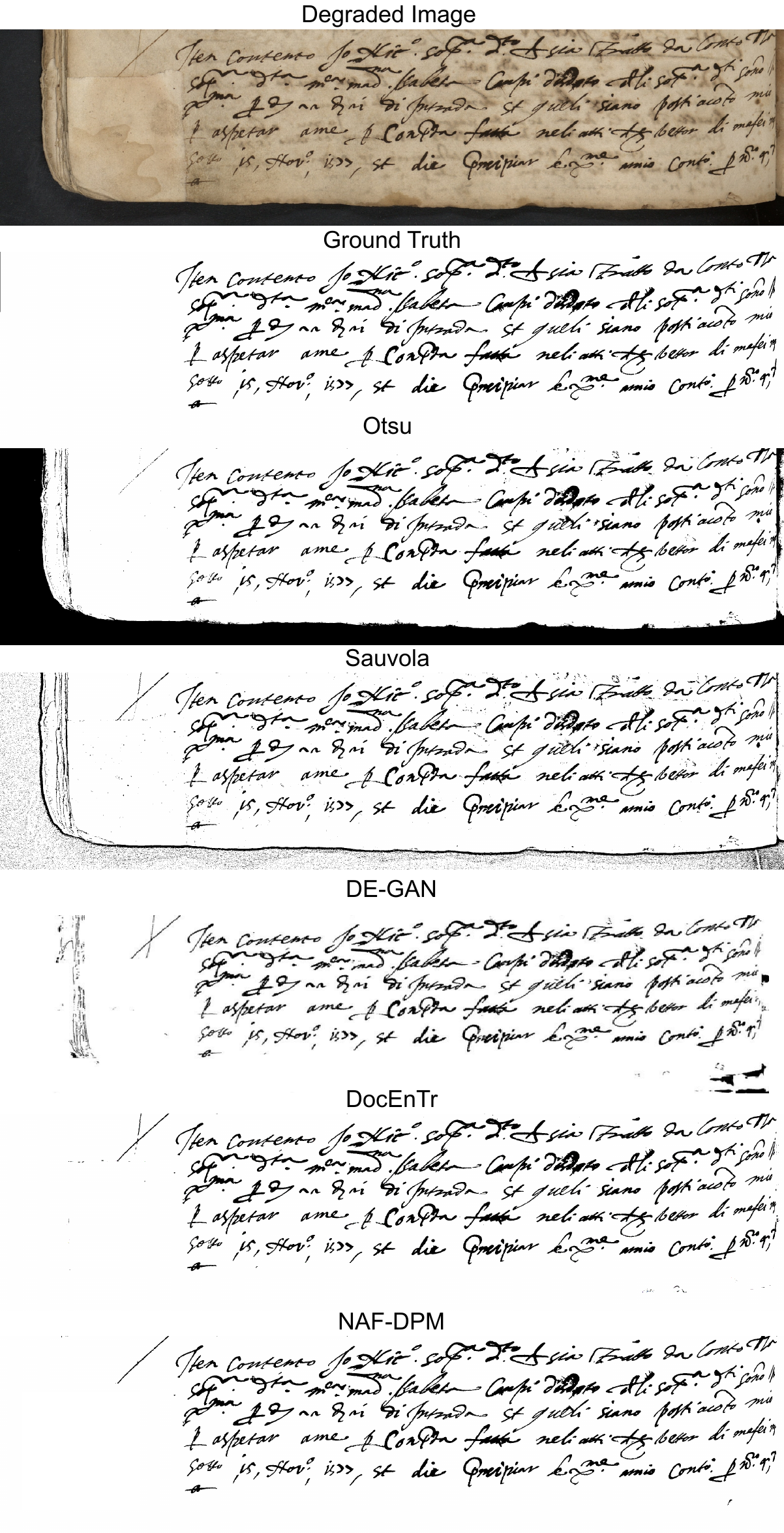}
\caption{Qualitative Result offered by different models on sample number 9 of H-DIBCO 2018 \cite{DIBCO2018}.}
\label{dibco2018 visual results}
\vspace{-0.25cm}
\end{figure}

We conduct also a qualitative analysis of our network and other different methods. In Figure~\ref{dibco2018 visual results} we compare an image from DIBCO2018 \cite{DIBCO2018} enhanced by our network and the same image processed by two basic thresholding techniques (Otsu \cite{otsu} and Sauvola \cite{SAUVOLA}), a conditional-GAN (DE-GAN \cite{DEGAN}) and a vision transformer (DocEnTr \cite{docentr}). 
It can be seen that dynamic threshold methods fail to precisely restore all portions of the image. Their small receptive field (25x25 sliding windows) does not allow them to get a complete view of the image. 
DEGAN \cite{DEGAN} succeeds in producing a good binarized version; however, there are some areas where the network demonstrates poor accuracy and poor generalization. 
The images enhanced by DocEnTR \cite{docentr} and by our method are almost overlapping and very similar to the ground truth.

\section{Conclusion}
In this work, we propose NAF-DPM, a framework that combines a convolutional neural network (CNN) and a diffusion probabilistic model (DPM) following the strategy ``predict and refine" with the final aim of restoring the original quality of degraded documents. As backbone network for the initial predictor and the diffusion model, we design an effective variant of a nonlinear activation-free network (NAFNet). To mitigate the slow sampling rate of diffusion models we employ an ODE fast solver which is able to converge in a maximum of 20 iterations. 
To better retain text characters, we introduced an additional module based on convolutional recurrent neural networks (CRNN) to simulate the behavior of an OCR system during the training phase. We test our approach on two use cases: document deblurring and document binarization. In both tasks, NAF-DPM produces high-quality samples with a few sampling iterations beating all previous powerful models on pixel-level and perceptual similarity metrics. It also significantly reduces character errors when commercial OCR systems transcribe the enhanced images.
Given the results obtained with NAF-DPM and considering the significant generalization power of diffusion models, possible future investigations might involve applying the method to other related tasks and developing a unified end-to-end framework for all the different document enhancement tasks. In particular, this latter work is important to reduce the number of different models and to decrease the training or finetuning phases needed to adapt the model to the various tasks.

\bibliographystyle{IEEEtran}

\bibliography{bibliography.bib}

\newpage

\vfill

\end{document}